\pdfoutput=1

\documentclass[11pt]{article}

\usepackage[]{acl}

\usepackage{times}
\usepackage{latexsym}
\usepackage{booktabs}
\usepackage{graphicx}
\usepackage{multirow}

\usepackage{graphicx}
\usepackage{amsmath}
\usepackage{amsfonts}
\usepackage{microtype}
\usepackage{booktabs} 
\usepackage{xspace}
\usepackage{subcaption}
\usepackage{bbm}
\usepackage{multirow}
\usepackage{enumitem}
\usepackage{xspace}
\newcommand{\model}{\textsc{MPR}\xspace}
\usepackage{cleveref}
\newcommand{\Scref}[1]{\S\ref{#1}}

\usepackage{kky}
\newcommand{\Dsrc}{\mathcal{D}_\text{src}}
\newcommand{\Dtgt}{\mathcal{D}_\text{tgt}}
\newcommand{\Dsyn}{\mathcal{D}_\text{syn}}
\newcommand{\Qtype}{\Qcal_\text{type}}
\newcommand{\Atype}{\Acal_\text{type}}

\usepackage[T1]{fontenc}

\usepackage[utf8]{inputenc}

\usepackage{microtype}

\newcommand{\draftonly}[1]{#1} 
\renewcommand{\draftonly}[1]{}

\usepackage[textsize=scriptsize]{todonotes}
	
\usepackage{color, colortbl}
\newcommand{\jhtodo}[1]{\draftonly{\todo{#1--\textsc{Junjie}}}}

\newcommand{\MPRgen}{MPR$_\text{gen} \ $}
\newcommand{\MPRdisc}{MPR$_\text{disc} $}
\newcommand{\MPRinit}{MPR$_\text{gen\_PM} \ $}
\newcommand{\MPRban}{MPR$_\text{disc\_BAN} \ $}

\title{Multimodal Prompt Retrieval for Generative Visual Question Answering}

\author{Timothy Ossowski$^1$, Junjie Hu$^{1,2}$\\
  $^1$Department of Computer Science, $^2$Department of Biostatistics and Medical Informatics\\
  University of Wisconsin, Madison, WI, USA\\
  \texttt{ossowski@wisc.edu}, \texttt{junjie.hu@wisc.edu}}

\begin{document}
\maketitle
\begin{abstract}
Recent years have witnessed impressive results of pre-trained vision-language models on knowledge-intensive tasks such as visual question answering (VQA). Despite the recent advances in VQA, existing methods mainly adopt a discriminative formulation that predicts answers within a pre-defined label set, leading to easy overfitting on low-resource domains with limited labeled data (e.g., medicine) and poor generalization under domain shift to another dataset. To tackle this limitation, we propose a novel generative model enhanced by multimodal prompt retrieval (\model) that integrates retrieved prompts and multimodal features to generate answers in free text. Our generative model enables rapid zero-shot dataset adaptation to unseen data distributions and open-set answer labels across datasets. Our experiments on medical VQA tasks show that \model outperforms its non-retrieval counterpart by up to 30\% accuracy points in 
a few-shot domain adaptation setting.\footnote{Our code is publicly available at \url{https://github.com/tossowski/MultimodalPromptRetrieval}}

\end{abstract}

\section{Introduction}
\label{sec:introduction}


Visual question answering (VQA) is a popular multimodal machine learning problem that challenges a model to answer a question posed about an image. As encouraged by recent advances in VQA, pioneering studies have investigated the application of VQA systems to low-resourced, knowledge-intensive domains such as medicine~\cite{Lin2021MedicalVQ}, where collecting domain-specific annotations is extremely costly and time-consuming. In particular, medical VQA has attracted increasing research interests~\cite{hasan2018overview}, with the target of supporting clinical decision-making such as acting as an auxiliary virtual ``diagnostic radiologist''~\cite{kovaleva-etal-2020-towards}.  

Despite recent progress in general VQA leveraging pre-training~\cite{chen2022murag}, retrieval \cite{wu2022multi}, or knowledge bases~\cite{narasimhan2018straight, shevchenko2021reasoning}, several challenges still exist for medical VQA. First, medical VQA systems still suffer from a stark lack of high-quality labeled data. As a result, it is essential to leverage domain adaptation techniques~\cite{zhou2019tua1} that rapidly adapt models trained from a similar dataset to a target dataset. Second, as the medical domain covers a wide variety of complex diseases, there exists a large distribution shift across medical datasets, significantly increasing the complexity of learning medical images and texts by deep neural models. However, many existing medical VQA methods mainly focus on in-domain evaluation, testing systems on a held-out test set under the same data distribution of the training data. Moreover, these methods often augment their model architecture with dataset-specific components such as an answer-type classifier \cite{zhan2020medical}, separate models for each question-type \cite{khare2021mmbert}, or specific pre-trained medical encoders \cite{moon2022multi}. These dataset-specific designs hinder the application of these medical VQA models across datasets in new domains. Furthermore, existing medical VQA approaches~\cite{tanwani2022repsnet, eslami2021does} often adopt a discriminative model architecture that predicts a fixed set of answers, limiting model generalization to different answer sets.

To tackle these challenges, we propose a domain-agnostic generative VQA model with multimodal prompt retrieval (MPR) that retrieves relevant VQA examples to construct multimodal prompts and generates arbitrary free text as the answers, removing the restriction of predicting a fixed label set. To augment the retrieval data, we also investigate a data augmentation strategy to create a synthetic medical VQA dataset from medical image-captioning data. Our experiments on two medical VQA datasets demonstrate the effective adaptation of our proposed method to a new target medical dataset, while also showing similar in-domain performance of our models to existing discriminative baselines. 
Our contributions are summarized below:
\begin{itemize}[leftmargin=10pt]\itemsep-0.2em
    \item We introduce a multimodal prompt retrieval module that improves VQA generalization across different data distributions even with noisy synthetic data and smaller retrieval datasets.
    \item We investigate a zero-shot dataset adaptation setting for medical VQA systems across datasets, encouraging future research on in-context prediction of VQA systems for dataset adaptation.
    \item We propose a novel prompt-based generative VQA model, which enables more flexible answer outputs and controllable generation guided by multimodal prompts.

\end{itemize}
\section{Preliminaries}
\label{sec:preliminary}
This section provides descriptions of the VQA task and the challenges faced in the medical domain.

\paragraph{Problem Setup}

Formally, given a VQA dataset of $n$ tuples $\mathcal{D} = \{(v_i,x_i,y_i)\}_{i=1}^n$, we aim to learn a model to predict an answer $y_i$ given a question $x_i$ and an image $v_i$. Conventionally, a model consists of an image and text encoder that maps the inputs $v_i$ and $x_i$ to the latent space of $\mathcal{V}$ and $\mathcal{X}$ respectively:
\begin{align}
    \mathbf{v}_i &= \text{ImgEncoder}(v_i) \in \mathcal{V} \\ 
    \mathbf{x}_i &= \text{TextEncoder}(x_i) \in \mathcal{X} 
\end{align}

Most prior works learn a discriminative model $f_\theta$ that directly estimates a probability distribution over all possible answers in a pre-defined label set, i.e., $f_\theta: \mathcal{V}, \mathcal{X} \rightarrow  \mathcal{Y}$. In contrast, we adopt a generative model $g_\phi$ that predicts words in a vocabulary $\Sigma$ to generate a varying length text string $z\in\Sigma^+$, and apply a deterministic function to map the answer string $z$ to the closest answer label $y\in\mathcal{Y}$. 


\paragraph{Dataset Adaptation:} We also focus on a dataset adaptation setting where a model is trained on a source labeled dataset $\mathcal{D}_\text{src}$ and further adapted to a target dataset $\mathcal{D}_\text{tgt}$ with a different label set, i.e., $\mathcal{Y}_\text{src}\neq \mathcal{Y}_\text{tgt}$. Thus, it is nontrivial for a discriminative model $f_\theta$ to perform adaptation over different label sets. For adaptation with generative models, we consider two strategies of using target labeled data for (a) in-context prediction without updating the source-trained models $g_\phi$ and (b) continued fine-tuning $g_\phi$. While our method focuses on in-context prediction (\Scref{sec:methods}), we also compare these two strategies in our experiments (\Scref{sec:analysis}).

\paragraph{Types of Medical VQA:}\label{para:type_desc} According to the annotations of popular medical VQA tasks (e.g., SLAKE \cite{liu2021slake} and VQA-RAD \cite{lau2018dataset}), there are two answer types $\Atype$: \textit{closed} answers where the set of possible answers are disclosed in a question (e.g., yes-no questions); and \textit{open} answers that can be free-form texts. Besides, there are multiple different question types $\Qtype$ such as organ, abnormality, or modality, indicating the medicinal category for which the question is intended. Prior medical VQA models \cite{zhan2020medical, eslami2021does} use a binary classifier to distinguish the two answer types based on questions and apply two discriminative models to predict answers, while we propose to predict both types of answers by a single generative model in this work.

\section{Methods}
\label{sec:methods}
\begin{figure*}[th!]
  \centering{\includegraphics[trim={0 0 0 20},width=0.9\textwidth]{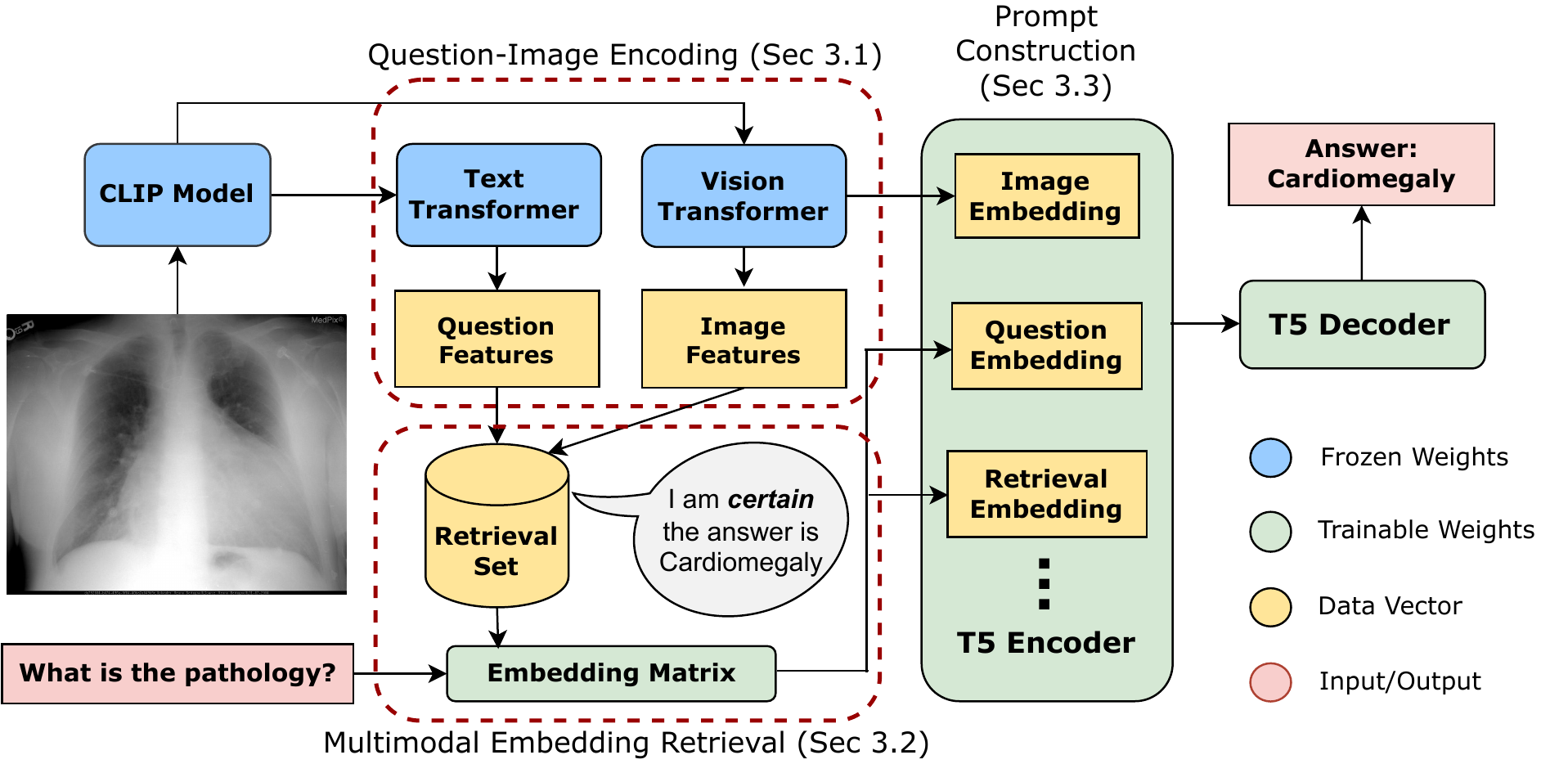}}
  \vspace{-3mm}
  \caption{\small An overview of multimodal prompt retrieval (MPR). There are three primary components of the prompt we use for the encoder indicated by the three yellow boxes contained in the T5 encoder block. These components can be optionally omitted or further extended with additional data.}
  \label{fig:prompt}
  \vspace{-3mm}
\end{figure*}

In this section, we start by introducing the text and image encoding for retrieval (\Scref{sec:method:encode_retrieval}), then describe the prompt construction from retrieval (\Scref{sec:method:prompt}), and prompt integration in our generative model (\Scref{sec:method:generative}). 

\paragraph{Overview:} For each $(v, x, y)\in\Dsrc$ during training, we propose to retrieve similar tuples from the training dataset $\Dsrc$, integrate the retrieved tuples for prediction, and update the model. We also assume to have access to a target labeled dataset $\Dtgt$ for dataset adaptation. Note that we mainly describe the in-context prediction using $\Dtgt$ here and leave the discussion of fine-tuning on $\Dtgt$ to the experiments. When predicting a target test example at test time, we directly apply our source-trained model to retrieve labeled tuples from $\Dtgt$ and perform prediction. The key insight is that even if the source-trained model is not directly trained on target data, the retrieved tuples may contain the correct answer to the given target question, potentially improving model predictions in the target dataset.

\subsection{Multimodal Prompt Encoding}
\label{sec:method:encode_retrieval}

For a VQA dataset, we can easily construct a mapping by using the image-question pair as the key and the answer as the value. Therefore we can use a multimodal encoder to encode the image-question pairs into multimodal features and perform K-Nearest Neighbors (KNN) search to find the most similar VQA tuples in the feature space.

\paragraph{Question-Image Encoding:} Before model training, we use a pre-trained CLIP model~\cite{radford2021learning} to encode image-question pairs in a retrieval dataset $\Rcal$, where $\Rcal=\Dsrc$ during training and $\Rcal=\Dtgt$ at testing. Specifically, we first preprocess each image by downsampling it to the $224 \times 224$ resolution and adopt CLIP's vision transformer~\cite{dosovitskiy2020image} to obtain image features $\vb_\texttt{CLS}$ of the image patch \texttt{[CLS]} token that summarizes the image content. Similarly, we process each question using CLIP's corresponding text transformer to obtain question features $\xb_\texttt{EOT}$ from the \texttt{[EOT]} token. 
These question and image features are concatenated to form a holistic vector representation $\pb=[\vb_\texttt{CLS}; \xb_\texttt{EOT}]$ of a question-image pair. These question-image vectors are paired along with the corresponding answers to construct the retrieval mapping set $\Mcal = \{(\pb_i, y_i)\}_{i=1}^m$ of $m$ items. 

\paragraph{Retrieval Set Augmentation on Image-Caption Data:} 
\label{sec:methods:construction}
As many VQA datasets in a low-resourced target domain (e.g., medicine) often contain a limited amount of labeled examples, we propose a data augmentation method to create a synthetic VQA set $\Dsyn$ from image-caption pairs and augment the retrieval set $\Rcal$. First, we determine a desired set of question types $\Qtype$ and answer types $\Atype$ described in \Scref{para:type_desc}. For each combination of question and answer types $t\in\Qtype\times\Atype$, we manually prepare a collection of question templates $\Tcal_t$
 along with a corresponding collection of keywords $\Wcal_t$. We then iterate through all the image-caption pairs and identify if the caption contains any keywords $w \in \Wcal_{t}$. If any keywords match, we create a question by sampling a template from $\Tcal_{t}$ uniformly at random and filling it with the matched keyword as the answer. Example templates from several question and answer types can be found in Appendix \ref{sec:templates}.
\subsection{Multimodal Embedding Retrieval}
\label{section:methods:retrieval}
To answer a question $x$ about an image $v$, we propose to retrieve its top-$k$ most similar examples from the retrieval mapping set $\Mcal$ (as constructed in \Scref{sec:method:encode_retrieval}). Specifically, we first encode the query question-image pair into an embedding $\pb$ by the CLIP model and compute the cosine similarity between the query embedding $\pb$ and each question-image embedding in $\Mcal$. Therefore, we can obtain the $k$ nearest neighbors of image-question pairs in $\Mcal$, denoted as $\Kcal = \{(\pb_i,y_i)\}_{i=1}^k $. Note that if the size of $\Mcal$ is large, KNN search can be implemented with efficient algorithms such as Maximum Inner Product Search~\cite{NIPS2014_310ce61c}. The retrieved pairs are used to construct the retrieval prompt (detailed in \Scref{sec:method:prompt}).

\subsection{Prompt Construction}
\label{sec:method:prompt}

Inspired by the prompt tuning method~\cite{lester-etal-2021-power} that appends several prompt embeddings to the original input before feeding to the transformer layers of the encoder, we propose to construct multimodal prompt embeddings to augment a question input, as shown in Figure~\ref{fig:prompt}. Specifically, given an image-question pair, the inputs to our model consist of three main components: image, question, and retrieval embeddings. Our model concatenates these embeddings as inputs to the subsequent stack of encoder layers in a T5 model~\cite{raffel2020exploring}. We begin the prompt with the image embedding, followed by the question and retrieval embeddings, leaving experimentation with alternative concatenation orders to Appendix \ref{sec:prompt_variations}.

\paragraph{Image Embedding:}
The image embedding is obtained using the same vision transformer of CLIP applied to construct the retrieval dataset. However, instead of using the \texttt{[CLS]} token which summarizes the image content, we use the intermediate output of the penultimate layer to obtain a collection of image token embeddings $\vb_p\in\RR^{l_v\times d}$, where $l_v$ denotes the number of image tokens. 

\paragraph{Question Embedding:}
To encode a question corresponding to an image, we use the embedding matrix of a pre-trained T5 encoder. Following the practice of T5, we include a short text snippet (e.g., ``Answer the abnormality question:'') at the beginning of the question to instruct the model to perform a QA task. The combined text is first tokenized according to T5's subword tokenization, followed by an embedding lookup to fetch the corresponding embedding vectors $\xb_q\in\RR^{l_q\times d}$ from T5's input embedding matrix. 

\paragraph{Retrieval Embedding:}
Based on the top-$k$ similar examples retrieved $\Kcal$, we define an ordered list of quantifier words $\Qcal=[q_1, \dots, q_M]$ (e.g., $[\text{very unlikely, ..., very likely, certainly}]$) and a text template $T_\text{prompt}$. We define a confidence score that counts the frequency of the retrieved answers in $\Kcal$, and then select the most frequent answer $y_r^*$ from $\Kcal$ in Eq.~(\ref{eq:retrieved_answer}). We then apply a threshold function to select an appropriate quantifier $q_r^*$ from $\Qcal$ based on the confidence score of $y_r^*$ by Eq.~(\ref{eq:quantifier}). 
\vspace{-1mm}
\begin{align}
    \label{eq:retrieved_answer} &\pb_r^*, y_r^* = \arg\max_{(\pb, y)~\in \Kcal} \text{Freq}(y, \Kcal)\\
    \label{eq:quantifier} &q_r^* = q_i, ~\text{if}~\frac{i-1}{M}\leq \frac{\text{Freq}(y_r^*, \Kcal)}{k} < \frac{i}{M}
\end{align}
\vspace{-1mm}
We then construct the retrieval prompt by filling in the template $T_\text{prompt}$ with the quantifier $q_r^*$ and the retrieved answer $y_r^*$. We detail example templates and prompt variants we explored in Appendix \ref{sec:templates}. The same pre-trained T5 model used for the question prompt is used to tokenize the retrieval prompt and obtain the retrieval embeddings $\xb_r\in\RR^{l_r\times d}$.





\subsection{Generative Visual Question Answering}
\label{sec:method:generative}

\paragraph{Encoder:} 
Following prompt construction, we obtain a combination of embeddings $[\vb_p;\xb_q; \xb_r]$ which is further fed as inputs to the transformer encoder layers of a pre-trained T5 model, and obtain contextualized representations of the combined sequence from the top encoder layer, where we denote as $\Xb=\text{Encoder}([\vb_p;\xb_q; \xb_r])$. In this work, we use a moderately sized model with around 60 million parameters, T5-small, and leave models with more parameters for future exploration.

\paragraph{Decoder:} While most prior works use a discriminative architecture for medical VQA, we experiment with a decoder to predict free-form text. A transformer decoder from T5 is used to predict words in the vocabulary autoregressively. As each answer label $y$ has a corresponding text string $z$ of varying length, we formulate the likelihood of an answer string $z$ given an image $\vb$ and a question $\xb$ by the following conditional probability:
\vspace{-1mm}
\begin{align}
    P_\text{gen}(z | \Xb) = \prod_{j=0}^{|z|}P_{\phi}(z_j | \Xb, z_{<j}).
\end{align}
\vspace{-1mm}
We finally optimize the generative model using a cross-entropy loss between the conditional probability $P_\text{gen}(z|\Xb)$ and the ground-truth answer string $z$ on the training dataset. 
This formulation allows for more flexible answers, which can easily change depending on the task, but may produce answers that are essentially the same with minor differences (e.g., extra whitespace, synonyms, etc.). To resolve these minor discrepancies, we utilize a simple string-matching heuristic that matches the longest continuous subsequence\footnote{\scriptsize \url{https://docs.python.org/3/library/difflib.html}} between the generated answer and the closest possible label in the answer label set. Thus, our final generative model predicts answers as follows:
\vspace{-1mm}
\begin{align}
    z^* &= \arg\max_{z} P_{gen}(z | \Xb) \\
    y^* &= \text{LongestCommonString}(z^*, \mathcal{Y}). 
\end{align}
\vspace{-1mm}
Compared to the exact match between the generated answer string $z^*$ and the ground-truth string $z$, we observe a 3-4\% improvement in accuracy when using this heuristic on the VQA-RAD dataset and a 1\% gain on the SLAKE dataset.


\section{Experimental Setup}
\label{sec:setup}

We perform our analysis on the VQA-RAD and SLAKE datasets, which are anonymous and preprocessed following prior works \cite{eslami2021does, zhan2020medical}. We use an AdamW optimizer with an initial learning rate $1e^{-4}$ for T5 finetuning. We use a $\texttt{ViT-B/32}$ architecture for our CLIP models, and T5-small for answer generation. The plateau learning rate scheduler is used to decay the learning rate by a factor of 10 if the validation loss does not decrease for 10 consecutive epochs.

All model training used a batch size of 16 and took 2-3 hours on average on a RTX 3090 GPU. All results were seeded with the best run of \citet{eslami2021does, zhan2020medical} for reproducibility.\footnote{The performance was similar across 5 different seeds for our best model, with a standard deviation of about 1.5\%.}

\subsection{Datasets}
\paragraph{SLAKE}
The SLAKE dataset comprises 642 images and over 14,000 VQA pairs in English and Chinese. We only use the English portion to match the language of the T5 pretraining corpus. We use the provided train, validation and test splits, corresponding to 4918, 1053, and 1061 QA pairs. SLAKE consists of 10 different question types.\footnote{Details in Table~\ref{tab:data_statistics}  and \ref{tab:question_types} in the appendix.} 

\paragraph{VQA-RAD}
VQA-RAD is a high-quality dataset consisting of 315 patient scans and 3515 questions. We use the train and eval splits provided with the original data, following prior works~\cite{tanwani2022repsnet, eslami2021does, nguyen2019overcoming}. VQA-RAD consists of 11 different question types. 

\paragraph{Radiology Objects in Context (ROCO)} The ROCO dataset~\cite{pelka2018radiology} has over 81,000 radiology image-caption pairs, making it a popular medical dataset for pretraining vision-language models \cite{pelka2018radiology}. Each image-caption pair also contains keywords and semantic types used by existing works for masked language modeling on salient spans \cite{khare2021mmbert}. 

\paragraph{Synthetic VQA Data}
Using the image-caption data from the ROCO dataset, we construct a large-scale synthetic VQA dataset consisting of over 50,000 question-answer pairs. Using our procedure (\Scref{sec:methods:construction}), we create question and keyword templates focusing on organ, modality, and plane questions.


\subsection{Training Settings}
\paragraph{Dataset Adaptation (DA):} To evaluate the generalization of VQA models across datasets, we examine a setting where we train a model on a source-labeled dataset and use it to answer questions from a different target dataset with access to a target dataset. We further compare models using the target labeled examples for (a) in-context prediction without updating source-trained models or (b) continued fine-tuning.

\paragraph{In-domain Evaluation (IDE):} In this setting, we adopt a standard split of each dataset into train/validation/test sets. We then train models on the train set, select the best checkpoints by the validation set, and evaluate models on the test set.

\subsection{Baselines}

\paragraph{Mixture of Enhanced Visual Features (MEVF)} \citet{nguyen2019overcoming} utilize model agnostic meta-learning (MAML) in conjunction with a convolutional denoising autoencoder (CDAE) to learn medical image latent feature representations. 


\paragraph{Question Answering with Conditional Reasoning (QCR)} \citet{zhan2020medical} introduce novel task-conditioned, open, and closed reasoning modules to distinguish between answer types and improve open question accuracy. 

\paragraph{PubMedCLIP} \citet{eslami2021does} utilize the ROCO dataset to finetune a general CLIP model on medical image-caption pairs. They modify existing architectures with the finetuned vision encoder to achieve improved results.

\paragraph{MMBERT} \citet{khare2021mmbert} introduces a BERT-based method that utilizes pretraining on the ROCO dataset with a masked language modeling objective. The model predicts answers by performing an average pooling on the last layer features followed by a linear classification layer.

\paragraph{\MPRdisc~(Ours):}
MPR$_\text{disc}$~refers to our discriminative variant by replacing a generative decoder with a prediction head to predict a finite set of answers. \MPRban uses a prediction head similar to \MPRdisc, but fuses the image and text features with a bilinear attention network~\cite{kim2018bilinear}.

\paragraph{\MPRgen (Ours):}
\MPRgen refers to our generative architecture which outputs flexible answers. \MPRinit has the same architecture as \MPRgen, but is initialized with a pre-trained checkpoint from PubMedCLIP~\cite{eslami2021does}.\footnote{\scriptsize \url{ https://github.com/sarahESL/PubMedCLIP}}

\begin{table*}[!htbp]
    \resizebox{\textwidth}{!}
    {
    \centering
    \scriptsize
        \begin{tabular}{cl|ccc|ccc}
        \hline
        \toprule[1pt]
        \multirow{2}{*}{Context} & \multirow{2}{*}{Method} & \multicolumn{3}{c|}{SLAKE $\rightarrow$ \text{VQA-RAD}}  &  \multicolumn{3}{c}{VQA-RAD $\rightarrow$ \text{SLAKE}} \\
         
         & & Open & Closed & Overall & Open & Closed & Overall \\
        \hline
        \multirow{2}{*}{Image and Question}
        &\MPRinit & 6.0 &  53.4 & 34.6 & 18.3 & 52.2 & 31.6  \\
        &\MPRgen & 4.9 & 52.0 & 33.3 & 16.9 &  46.4 & 28.5\\
        \hline
         \multirow{2}{*}{Image, Question, and Retrieval}
        &\MPRinit   & 42.9 & 76.2 & 63.0 &  45.1 & 67.3 & 53.8\\
        &\MPRgen & 41.8 & 74.4 & 61.4 & 38.4 & 57.7 & 46.0\\
        \bottomrule[1pt]
        \end{tabular}
        }
    \vspace{-3mm}
    \caption{\small Performances of our generative prompting method in a domain adaptation setting with different levels of context. When provided with retrieval context, the models query for $k = 1$ relevant image-question pairs. }
    \label{tab:cde_overall}

\end{table*}

\begin{figure*}[!t]

  \includegraphics[width=0.5\textwidth]{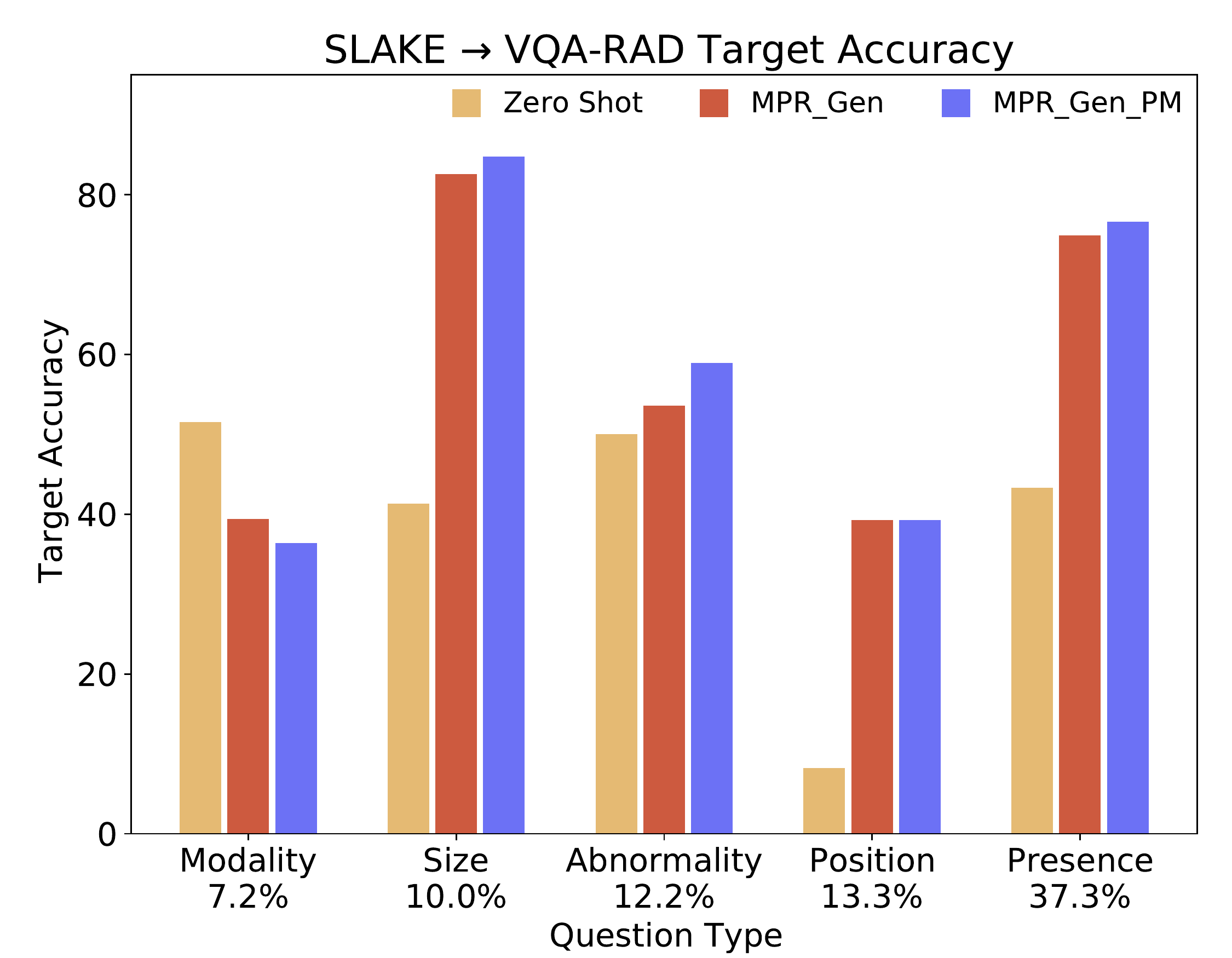}
  \includegraphics[width=0.5\textwidth]{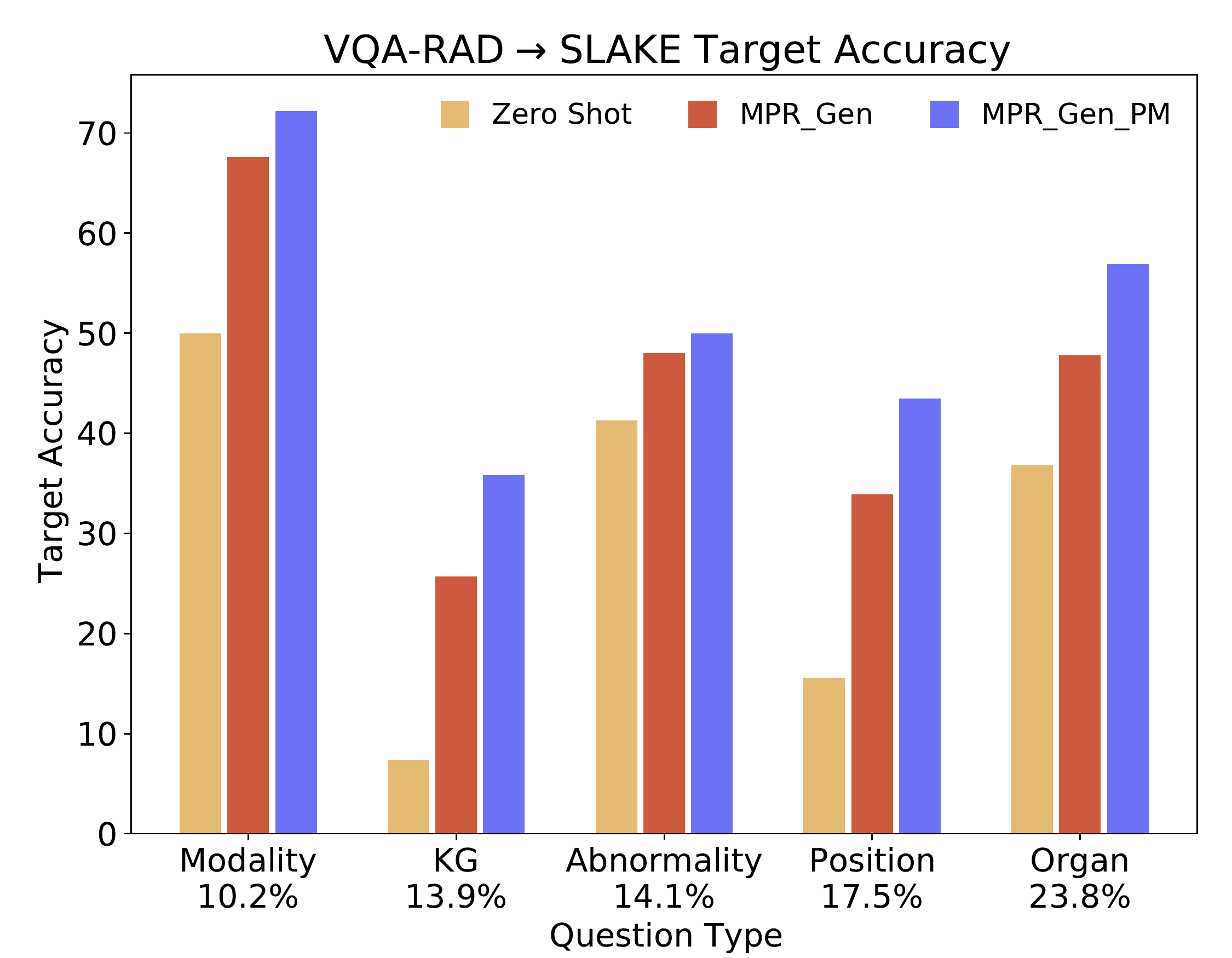}
  
  \vspace{-2mm}
  \caption{Domain adapation accuracy for the top 5 most common question types in the SLAKE and VQA-RAD datasets. Percentages on the x-axis indicate each question type's proportion of the dataset. \MPRgen and \MPRinit use retrieval and query for $k = 1$ relevant image-question pairs. Retrieval-based models outperform our zero-shot baseline, and initializing MPR with a PubMedCLIP checkpoint helps. }
  \label{fig:DA-finegrained}
  \vspace{-3mm}
\end{figure*}

\section{Results and Analysis}
\label{sec:analysis}
This section describes the results of our main experiments (\Scref{sec:main}) and fine-grained analysis thereafter. 

\subsection{In-context Prediction for Adaptation}
\label{sec:main}
First, we evaluate our proposed method's generalization capability of in-context predictions. We define a $k$-shot setting where our model retrieves Top-$k$ similar image-question pairs from the retrieval set. We compare the performance of the $k = 1$ setting with zero-shot MPR (i.e., MPR w/o retrieval) on two medical domain adaptation tasks. 

\paragraph{Overall Accuracy:} 
Table \ref{tab:cde_overall} compares the performances of our generative models under domain shift. Most notably, allowing the models to access a retrieval set universally improves performance, especially on questions with open answers. We also demonstrate that initializing our model with a PubMedCLIP pre-trained checkpoint results in higher accuracy than a general CLIP checkpoint. As the other discrinimative baselines can only predict a fixed set of answers, they cannot perform adaptation over different answer sets. We only compare them for our in-domain analysis (\Scref{sec:in-domain}).

\vspace{-2mm}
\paragraph{Fine-grained Accuracy over QA Types:}
Figure \ref{fig:DA-finegrained} summarizes our model performances across individual QA types in a domain adaptation setting. We find that zero-shot MPR struggles with question types that require logical reasoning, such as Knowledge Graph (KG) or Position questions, while in-context retrieval increases model performance significantly in these question types. Using a PubMedCLIP vision encoder further increases accuracy for these challenging question types.

\vspace{-3mm}
\subsection{Retrieval Sets for In-context Prediction}
\label{sec:synthetic_retrieval}

\begin{table}[htb!]    

    \renewrobustcmd{\bfseries}{\fontseries{b}\selectfont}

    \centering
    \resizebox{\linewidth}{!}
    {
        \begin{tabular}{c@{\hspace{1.5\tabcolsep}}c@{\hspace{1.5\tabcolsep}}c@{\hspace{1.5\tabcolsep}}c@{\hspace{1.5\tabcolsep}}c}
        \toprule
        \toprule[1pt]
         Source $\rightarrow$ Target & Retrieval Set & Open & Closed & Overall \\

        \midrule
        \multirow{4}{*}{SLAKE $\rightarrow$ VQA-RAD} & None (Zero-shot) & 6.0 & 53.4 & 34.6 \\
        & Synthetic & 11.5 & 49.8 & 34.6\\
        & VQA-RAD   & 42.9 & 76.2 & 63.0\\
        & VQA-RAD + Synthetic & 44.5 & 76.5 & 63.8  \\
        \midrule
        \multirow{4}{*}{VQA-RAD $\rightarrow$ SLAKE} & None (Zero-shot)  & 16.9 & 46.4 & 28.5 \\
        & Synthetic  & 18.3 & 50.2 & 30.8\\
        & SLAKE & 45.1 & 67.3 & 53.8 \\
        & SLAKE + Synthetic & 45.1 & 67.3 & 53.8 \\
        \bottomrule[1pt]
        \end{tabular}
    }
    \vspace{-2mm}
    \caption{\small Results of zero-/few-shot in-context prediction for domain adaptation with varying degrees of retrieval dataset access. We use \MPRinit with $k=1$ for all settings except $k=50$ for the noisy synthetic dataset.}
    \vspace{-4mm}
    \label{tab:zeroshot}

\end{table}

We also examine the effect of using different datasets for retrieval. Table \ref{tab:zeroshot} illustrates the zero-shot/few-shot accuracies when applying a source model to a target dataset with different retrieval datasets. Increasing the retrieval dataset's quality improves the model's adaptation capability to new questions. Without any retrieval, open question accuracy is as low as 6\%. Providing access to a noisy synthetic retrieval dataset improves open question performance. Using a higher quality in-domain retrieval set further enhances performance in all categories, achieving over 30\% improvement in open question accuracy compared to the zero-shot baselines. Combining in-domain retrieval data with noisy synthetic data further boosts accuracy in all three accuracy categories on VQA-RAD. However, we observed no further improvement when combining the synthetic and SLAKE datasets. With a manual investigation, we find that questions in SLAKE have much simpler synthetic variants than those in VQA-RAD. Therefore, SLAKE already provides the most similar examples during retrieval, and additional synthetic data provides minimal gains.

\begin{figure}[!t]
   \hspace*{0.2cm}\includegraphics[trim={0 20 0 40},width=0.4\textwidth]{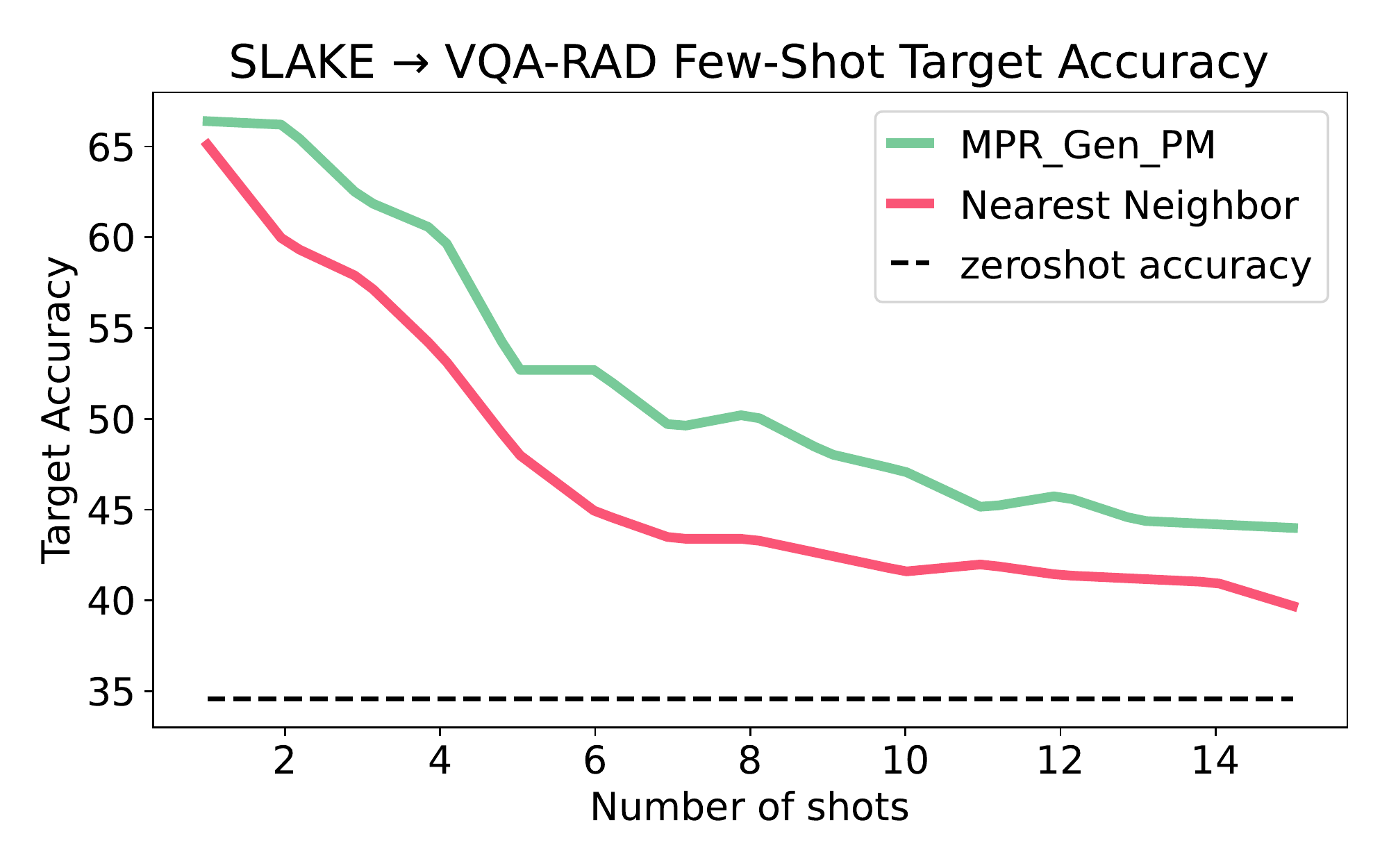}
  \vspace{-2mm}
   \captionsetup{belowskip=-10pt}
  \caption{\small A generative model \MPRinit trained on the SLAKE dataset is evaluated on the VQA-RAD dataset over different numbers of few-shot retrievals. }
  \label{fig:fewshot}
\end{figure}

\vspace{-1mm}

\subsection{How Many Shots are Needed?}
\label{sec:fewshots}
For adaptation at test time, we investigate the effect of varying the number of retrieved image-question pairs from the target dataset for constructing the retrieval prompts in Figure \ref{fig:fewshot}. Regardless of the number of pairs retrieved, the overall target accuracy of \MPRgen is always above the none-retrieval baseline (i.e., zero-shot). We hypothesize that accuracy peaks when $k = 1$ and stabilizes as $k$ increases due to the small dataset size. \MPRgen outperforms a purely nearest neighbor-based approach when testing on the VQA-RAD dataset. However, on a syntactically simpler dataset (i.e., SLAKE), we also find that a nearest neighbor-based classifier can achieve higher accuracy than our model. 

\vspace{-1mm}

\subsection{In-context Prediction vs Finetuning}
\begin{figure}[!t]


\captionsetup{belowskip=-10pt}

    \vspace{-5mm}
  \hspace*{0.2cm}\includegraphics[width=0.45\textwidth]{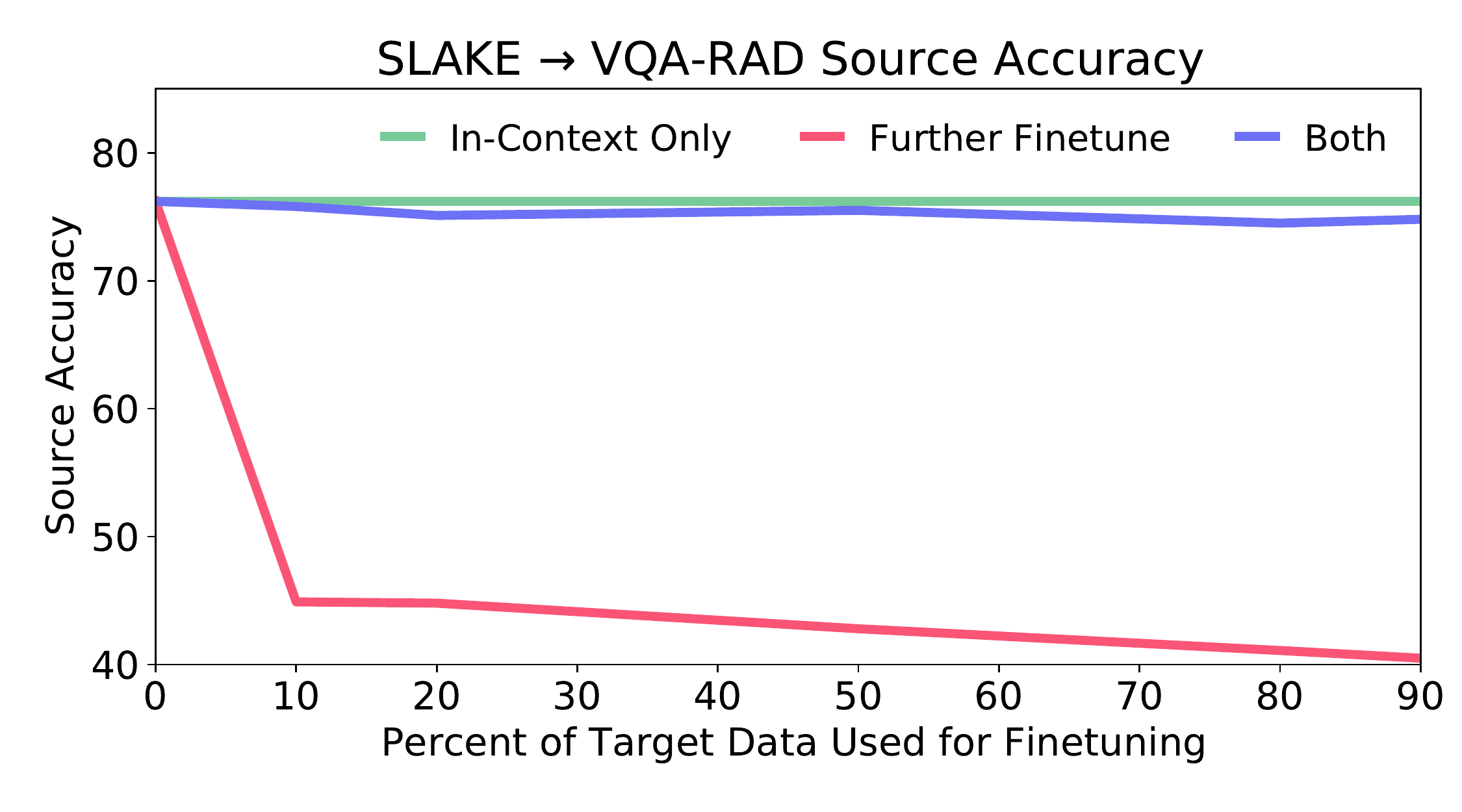}
  \hspace*{0.2cm}\includegraphics[width=0.45\textwidth]{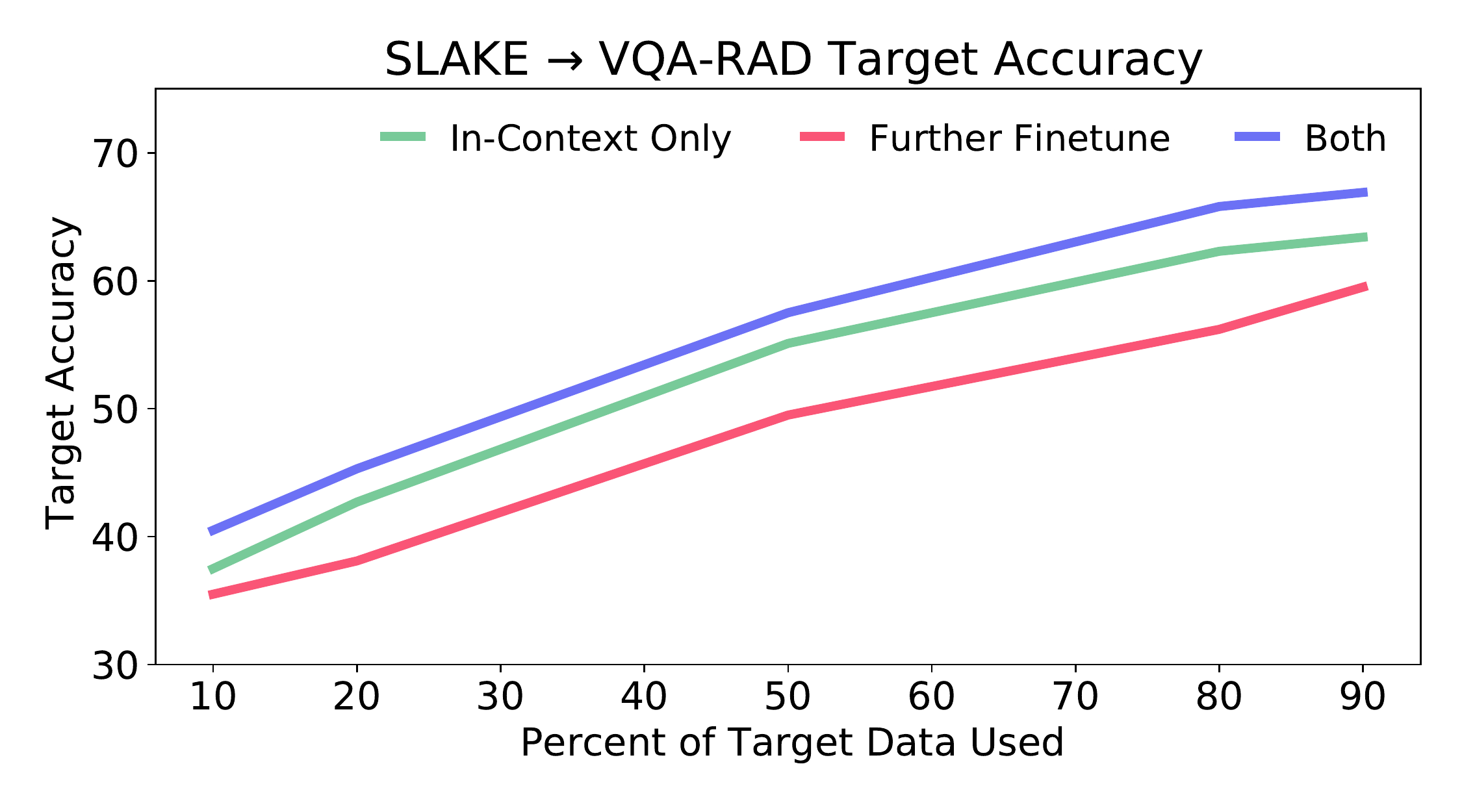}

    \vspace{-3mm}
  
  \caption{\small We vary the finetuning dataset size and observe the accuracy of finetuning compared to in-context prediction. All models start from a \MPRinit checkpoint and are either further finetuned (red + blue) or kept the same (green). }
  
  \label{fig:further-finetune}
\end{figure}

While further finetuning neural models on the target dataset often successfully learns to adapt to the new distribution, this technique often results in catastrophic forgetting \cite{thompson2019overcoming}. Figure \ref{fig:further-finetune} shows our experiments with further finetuning a source-trained model on a target dataset. First, we initialize three models with a \MPRinit checkpoint trained on SLAKE and adapt them to VQA-RAD. The first model is frozen, only using in-context prediction with retrieved target data (green). Another model is further finetuned on the target data without in-context prediction (red). The last model uses fine-tuning first and then does in-context predictions with retrieval (blue). 

Several findings can be observed. First, we find that in-context prediction with \MPRinit can mitigate the forgetting issue and improve cross-dataset adaptation. Second, when target data is scarce, in-context prediction outperforms further finetuning. Although the finetuned model achieved higher test accuracy when using all the target data, it suffered significant performance loss in its original domain. Lastly, combining in-context prediction with further finetuning eliminates most of this forgetting with minimal target domain performance loss.

\begin{table*}[!t]

    \resizebox{\textwidth}{!}
    {
    \centering
    \scriptsize
        \begin{tabular}{cc|ccc|ccc}
        \hline
        \toprule[1pt]

         \multirow{2}{*}{Context} & \multirow{2}{*}{Method} & \multicolumn{3}{c|}{SLAKE}  &  \multicolumn{3}{c}{VQA-RAD} \\
         &  & Open & Closed & Overall & Open & Closed & Overall \\
        \hline
        \multirow{2}{*}{Question Only} & \MPRgen  & 45.6 & 68.3 & 54.9 & 22.5 & 63.5 & 50.3\\

        & \MPRdisc & 48.5 &  66.6 & 55.6 & 38.5 & 72.6 & 59.0 \\
        \hline
        \multirow{8}{*}{Image and Question}
        &\MPRgen & 71.5 &  76.7 & 73.5 & 57.7 & 77.6 & 69.7  \\
        &\MPRinit & 74.1 & 82.2 & 77.3 & 62.6 & 78.3 & 72.1\\
        &\MPRdisc & 78.3 & \bfseries 84.9 & \bfseries 80.9 & 57.7 & 76.2 & 68.8 \\
        &\MPRban & 76.0 & 79.8 &  77.5 & 60.4 & \bfseries 81.6 & \bfseries 73.2\\
        & PubMedCLIP \cite{eslami2021does}  & \bfseries 78.4 & 82.5 &  80.1 & 60.1 & 80.0 & 72.1\\
        & MMBert \cite{khare2021mmbert}  & - & - &  - & \bfseries 63.1 & 77.9 & 72.0\\
        & QCR \cite{zhan2020medical} & - & - & - & 60.0 & 79.3 & 71.6 \\
        & MEVF \cite{nguyen2019overcoming} & - & - & - & 43.9 & 75.1 & 62.7\\
        \hline
         \multirow{4}{*}{Image, Question, and Retrieval}
        &\MPRgen   & 73.0 & 79.8 & 75.7 & 57.7 & 77.3 & 69.5\\
        &\MPRinit & 73.5 & 80.5 & 76.2 & 60.4 & 80.9 & 72.8\\
        & \MPRdisc & 75.0 & 81.0 & 77.4 & 51.6 & 78.0 & 67.5 \\
        & \MPRban & 77.5 & 82.2 & 79.4 & 62.6 & 80.1 & 73.2\\
        \bottomrule[1pt]
        \end{tabular}
        }
    \vspace{-2mm}
    \caption{\small Performances of our prompting method with different levels of context provided to the model. In most cases, our model is competitive with state-of-the-art methods even in the generative based cases. Retrieval context is provided by querying for the 15 most relevant image-question pairs. Bold values indicate the maximum in each column.}
    \label{tab:overall}
    \vspace{-3mm}
\end{table*}
\vspace{-1mm}
\subsection{In-domain Evaluation}
\label{sec:in-domain}

\paragraph{Overall Accuracy} We also compare our proposed model with existing models for the in-domain setting on SLAKE and VQA-RAD. We highlight the overall, open, and closed test accuracy for each dataset. We also evaluate our method with three contexts to analyze the effect of each component of our prompting method in Table \ref{tab:overall}.

As expected, the model variants perform worse when we only provide questions as inputs. Under the same setting where both the question and image features are provided, our generative model is competitive with the state-of-the-art discriminative models. Besides, we also find that using an in-domain dataset for retrieval does not provide performance gains, indicating that models can easily fit a small in-domain dataset, and retrieving prompts from the same training set does not provide extra useful information.

\paragraph{Finegrained Accuracy}
Figure \ref{fig:finegrained} in Appendix also shows the in-domain performance of our model variants across different question types for both datasets. The results indicate that all models generally struggle with questions requiring more complex reasoning, such as Position, Abnormality, and Knowledge Graph (KG) questions.

\section{Related Work}
\label{sec:related}

\paragraph{Retrieval-Based VQA}
Retrieval-based methods typically combine parametric models with non-parametric external memory for prediction. This idea first surfaces in KNN-LMs \cite{khandelwal2019generalization}, which utilizes a static retrieval data store to help language models adapt rapidly to new domains without further training. \citet{guu2020retrieval} extends this idea by introducing a parametric retriever that learns to attend to relevant documents during training. Recently, \citet{gao2022transform} summarizes visual information into natural language to use as a query for dense passage retrieval. The retrieved passages allow for the VQA model to outperform existing works, especially on questions which require outside knowledge. \citet{lin2022retrieval} consider training the retriever in an end-to-end manner similar to \citet{lewis2020retrieval} and find that this results in higher answer quality and lower computational training cost.


Different from these methods, we propose to construct multimodal prompts from retrieval to perform zeroshot dataset adaptation. While dataset adaptation of VQA models has been investigated in \citet{agrawal2022rethinking}, we focus on the effect of retrieval on generalization capability.
\jhtodo{What're the main differences of these prior works compared to ours?}

\paragraph{Generative QA}
Generative QA models focus on predicting answers autoregressively based on the input question. In this setting, the model may either generate the response based solely on model parameters (closed book) \cite{khashabi2020unifiedqa, roberts2020much} or rely on additional retrieved contexts (open book) \cite{karpukhin2020dense, lewis2020retrieval}. Our prompt construction method is inspired by the retrieval augmented generator (RAG) model \cite{lewis2020retrieval}, which retrieves relevant documents to answer questions. Instead of retrieving documents exclusively, we identify suitable image-question pairs to perform VQA.

\paragraph{VQA}
First introduced by \citet{antol2015vqa}, most VQA systems learn a joint embedding space for images and text to answer questions \cite{malinowski2015ask, gao2015you}. These approaches combine image and text features through either bilinear pooling or attention-based mechanisms \cite{yang2016stacked, lu2016hierarchical, anderson2018bottom, guo2021bilinear}. To help models understand the relationships between objects in an image, graph convolutional neural networks were introduced for VQA \cite{norcliffe2018learning, li2019relation}. Current methods often combine supplemental knowledge with fusion-based approaches to achieve state-of-the-art performance \cite{shevchenko2021reasoning, marino2021krisp, wu2022multi, chen2022murag}. We take a similar approach by using supplementary knowledge to construct context-aware prompts.
\section{Conclusion}
\label{sec:conclusion}

In this work, we propose a flexible prompt-based method for VQA in the medical domain. While our approach is designed for low-resource domains, the generative architecture of our model, in combination with a retrieval component, enables generalization to other fields. Our results are on par with state-of-the-art accuracies on the SLAKE and VQA-RAD datasets and show promising zero-shot and few-shot transfer results across different medical datasets. We hope these results can offer a baseline to compare with future work on knowledge-intensive and reasoning tasks.
\section{Limitations}
\label{sec:limitations}

When evaluating our model in a cross-dataset adaptation setting, our experiments indicate the importance of using a retrieval dataset. It is challenging to procure high-quality and volume retrieval datasets, especially in low-resource domains such as the medical field. Fortunately, the VQA-RAD and SLAKE datasets we evaluate on contain professionally annotated medical images. We also overcome the lack of data by creating a synthetic dataset from the medical ROCO image-captioning dataset.

Additionally, our model struggles with questions requiring multi-step reasoning, such as knowledge graph, abnormality, and position questions. Although performances in these question types are not far below the overall accuracy, future work may consider supplementary knowledge-based retrieval to assist in these challenging question types.

\section{Ethics Statement}
\label{sec:ethics}

Although medical VQA provides exciting opportunities for future AI-assisted clinical diagnostic tools, there are several ethical challenges associated with these approaches.

\paragraph{Patient Safety and Model Transparency}
Since the model decision process for deep learning models is difficult to understand, these models should only be used as an auxiliary tool. This obscure decision process is crucial to clarify in the medical domain, in which a poor diagnosis or choice of treatment can significantly affect patient lives. For example, medical experts found that cancer treatment recommendation software often gave unsafe or incorrect treatment advice in a recent study \cite{ross2018ibm}.

\paragraph{Dataset Biases}
The fairness of medical AI systems depends on the distribution of people in its training dataset. To ensure that AI algorithms display fairness to all races, genders, and ethnic groups, practitioners should verify that the training dataset contains an equal representation of all groups. Before deploying our architecture or other deep learning-based models to a clinical setting, practitioners should ensure that their patient's background is adequately represented in the training data.


\bibliography{anthology,custom}
\bibliographystyle{acl_natbib}

\clearpage
\appendix
\onecolumn
\section*{Appendix}

\label{sec:appendix}
\begin{table*}[h]
    \footnotesize
    \renewrobustcmd{\bfseries}{\fontseries{b}\selectfont}

    \centering
    \resizebox{\linewidth}{!}
    {
        \begin{tabular}{l@{\hskip4pt}l@{\hskip4pt}l@{\hskip4pt}l}
        \hline
        \toprule[1pt]
         Q Type & A Type & Question Templates \& Answer Templates ($\Tcal_t$) & Answer Templates\\

        
        \hline

        Organ & Open & Q: What part of the body is being imaged? What is the organ shown in this image? ... & A:\{Brain, Chest, ...\}\\
        Organ & Closed & Q: Does the picture contain \{\}? Is this a study of the \{\}? ... & A:\{Yes,No\}\\
        Organ System & Open & Q: What organ system is pictured? What system is this pathology in? ... & A:\{Respiratory System, Cardiovascular System, ...\}\\
        Organ System & Closed & Q: Is this an image of the \{\}? Is the \{\} shown? ... & A:\{Yes,No\}\\
        Modality & Open & Q: What kind of scan is this? How was this image taken? ... & A:\{MRI, X-ray, ...\}\\
        Modality & Closed & Q: Is this a \{\}? Is the image a \{\}? ... & A:\{Yes,No\}\\
        Plane & Open & Q: What image plane is this? How is the image oriented? ... & A:\{Axial, Coronal, ...\}\\
        Plane & Closed & Q: Is this a \{\} plane? Is the image a \{\} section? ... & A:\{Yes,No\}\\

        \bottomrule[1pt]
        \end{tabular}
    }

    \caption{Example templates for different question categories and question types.}
    \label{tab:templates}

\end{table*}
\begin{table*}[h]
    \footnotesize
    \renewrobustcmd{\bfseries}{\fontseries{b}\selectfont}

    \centering
    \resizebox{\linewidth}{!}
    {
        \begin{tabular}{ll}
        \hline
        \toprule[1pt]
         Q\&A Types ($t\in\Qtype\times\Atype$) & Keywords ($\Wcal_t$)\\

        
        \hline

        Organ \& Open & Heart, Lungs, Lung, Liver, Breasts, Chest, Cardiovascular System, Respiratory System ...\\
        Plane \& Open & Axial, Coronal, Supratentorial, Posteroanterior ... \\
        Modality \& Open & MRI, T1, T2, CT, X-ray, Ultrasound, Flair ... \\

        \bottomrule[1pt]
        \end{tabular}
    }

    \caption{Example keywords for different question types.}
    \label{tab:keywords}

\end{table*}
\section{Templates}
\label{sec:templates}
We use the question and keyword templates in Tables \ref{tab:templates} and \ref{tab:keywords} to construct a synthetic retrieval set. For open questions, the question template is static. However, the answer to open questions may be any of the keywords $w \in \Wcal_t$. Closed question templates have a slot that is filled in by one of the keywords $w \in \Wcal_t$, and the answer to these questions is either yes or no:

\begin{table*}[!h]  
    \footnotesize
    \renewrobustcmd{\bfseries}{\fontseries{b}\selectfont}

    \centering
    \resizebox{\linewidth}{!}
    {

        \begin{tabular}{@{\hskip1pt}c@{\hskip5pt}c@{\hskip5pt}c@{\hskip5pt}c@{\hskip5pt}c@{\hskip5pt}c@{\hskip1pt}}
        \hline
        \toprule[1pt]
         Prompt Construction Order & Prompt Template ($T_\text{prompt}$) & Possible Quantifiers ($\Qcal$) & Open & Closed & Overall \\

        
        \hline

        Question, Retrieval, Image & I believe the answer is \{quantifier\} \{answer\} & very unlikely, unlikely, maybe, likely, very likely, certainly & 39.6 & 65.0 & 54.9 \\
        
        Image, Retrieval, Question & I believe the answer is \{quantifier\} \{answer\} & very unlikely, unlikely, maybe, likely, very likely, certainly & 39.0 & 65.3 & 54.9 \\
        Image, Question, Retrieval & \{answer\} is \{quantifier\} the answer & very unlikely, unlikely, maybe, likely, very likely, certainly & 37.9 & 68.6 & 56.4 \\
        Image, Question, Retrieval & I believe the answer is \{quantifier\} \{answer\} & very unlikely, unlikely, maybe, likely, very likely, certainly & 39.6 & 65.3 & 55.1 \\
        \bottomrule[1pt]
        \end{tabular}
    }

    \caption{Using different variants of our prompt results in similar performances. This table shows the open, closed, and overall accuracy of \MPRinit in a domain adaptation setting from SLAKE to VQA-RAD, retrieving k=1 nearest question-image pairs.}
    \label{tab:prompt_variations}

\end{table*}
\section{Prompt Variations}
\label{sec:prompt_variations}
During the prompt construction process, we experiment with different prompt ordering and wording of retrieval prompts. We illustrate different template wording in the Prompt Template column of Table \ref{tab:prompt_variations}. Each prompt contains a quantifier that is filled in with an expression ranging from ``very unlikely" to ``certainly" based on the confidence score of $y^*$, described in Section \ref{section:methods:retrieval}. We found that performance does not significantly change when changing these aspects of the prompt, and we ultimately decided to use settings in the last row of the table.

\section{Dataset Information}
Tables \ref{tab:data_statistics} and \ref{tab:question_types} report descriptive statistics about the datasets used in our experiments. Although the synthetic data contains more question-answer pairs than SLAKE and VQA-RAD, it has noisier labels and more limited question types. SLAKE and VQA-RAD have larger question-answer diversity and share several question types, such as Organ, Position, and Abnormality questions. 
\begin{table*}[!h]  
    \tiny
    \renewrobustcmd{\bfseries}{\fontseries{b}\selectfont}

    \centering
    \resizebox{\linewidth}{!}
    {
        \begin{tabular}{ccccc}
        \hline
        \toprule[1pt]
         Dataset & Train Split & Validation Split & Test Split & Number of Question Types \\

        
        \hline

        SLAKE & 4918 & 1053 & 1061 & 10 \\
        VQA-RAD & 3064 & - & 451 & 11 \\
        ROCO (image-caption only) & 65460 & 8183 & 8182 & - \\
        Synthetic & 56526 & - & - & 3 \\

        \bottomrule[1pt]
        \end{tabular}
    }
    \vspace{-2mm}
    \caption{Statistics about the datasets in our experiments. The synthetic data contains three different question types because other types require domain-specific knowledge (e.g. abnormality, knowledge graph, etc).}
    \vspace{2mm}
    \label{tab:data_statistics}

\end{table*}
\begin{table*}[!h]  
    \footnotesize
    \renewrobustcmd{\bfseries}{\fontseries{b}\selectfont}

    \centering
    \resizebox{\linewidth}{!}
    {
        \begin{tabular}{cc|cc|cc}
        
        \toprule[1pt]

        \multicolumn{2}{c|}{SLAKE} & \multicolumn{2}{c|}{VQA-RAD} & \multicolumn{2}{c}{Synthetic}\\
         Question Type & Percentage of Test Data & Question Type & Percentage of Test Data & Question Type & Percentage of Data\\
         \hline
         Shape & 0.66 &  Color & 0.87 &  Modality & 30.61\\
         Color & 3.20 &  Quantity/Counting & 1.31 &  Plane & 30.65\\
         Quantity & 4.90 &  Organ & 2.18 &  Organ & 38.74\\
         Plane & 5.47 &  Attribute & 4.36 &  - & -\\
         Size & 6.13 &  Other & 5.66 &  - & -\\
         Modality & 10.18 &  Plane & 5.66 &  - & -\\
         Knowledge Graph & 13.95 &  Modality & 7.19 &  - & -\\
         Abnormality & 14.14 &  Size & 10.02 &  - & -\\
         Position & 17.53 &  Abnormality & 12.20 & - & -\\
         Organ & 23.85 &  Position & 13.29 &  - &-\\
         - & - &          Presence &  37.25 &  - & -\\

        
        \hline

        \bottomrule[1pt]
        \end{tabular}
    }
    \vspace{-2mm}
    \caption{Composition of the datasets in our experiments. We use the English portion of the SLAKE dataset.}

    \label{tab:question_types}

\end{table*}

\begin{figure*}[!h]
\centering
\includegraphics[width=0.9\textwidth]{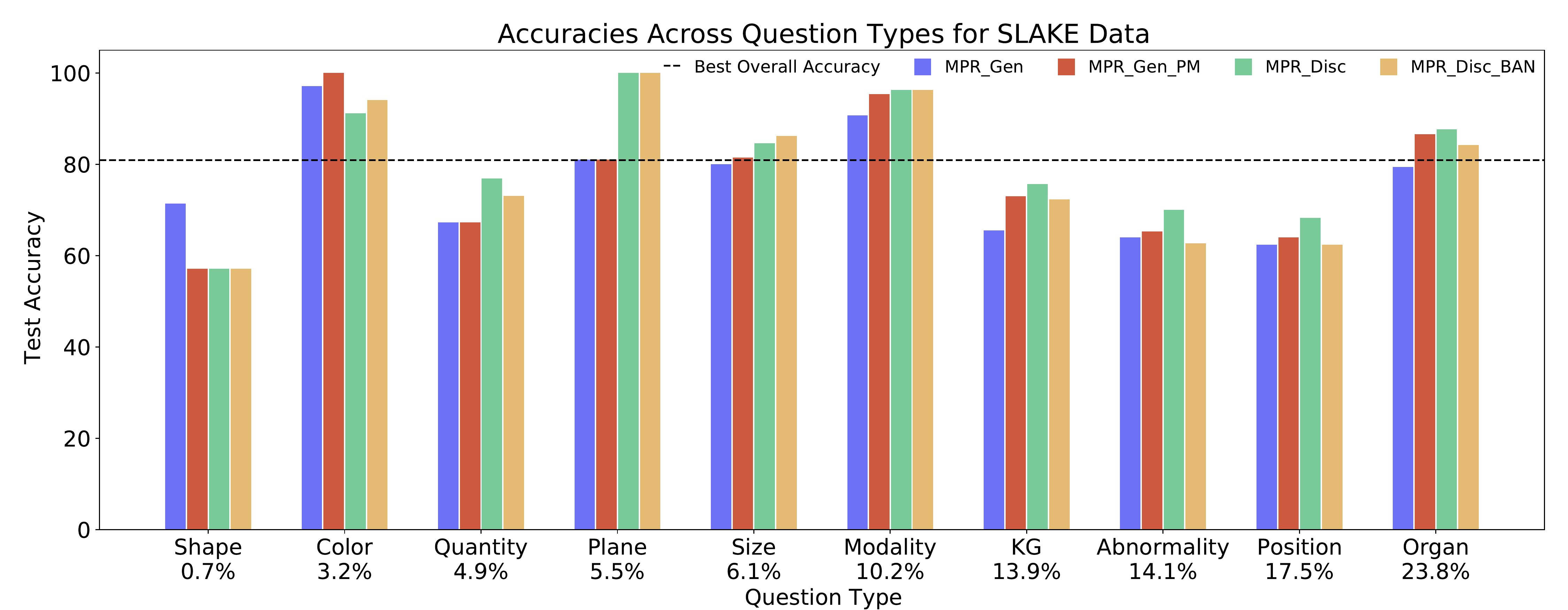}
\vspace{2mm}
\includegraphics[width=0.9\textwidth]{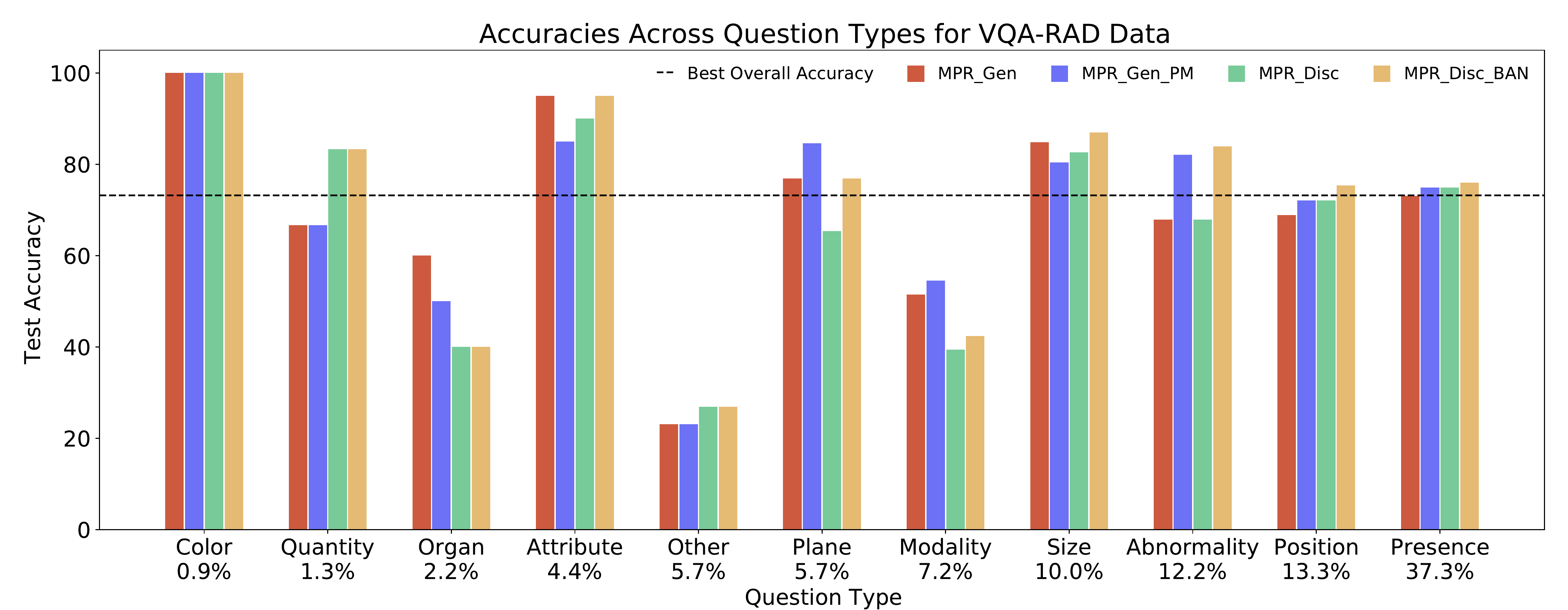}
  \vspace{-2mm}
  \caption{Accuracy for each question type for the SLAKE and VQA-RAD test datasets. Percentages on the x-axis indicate each question type's proportion of the dataset.}
  \label{fig:finegrained}
\end{figure*}

Figure \ref{fig:finegrained} displays our model's in-domain accuracies on SLAKE and VQA-RAD. The models perform best on Color, Attribute, and Size questions in VQA-RAD. The discriminative variants have better accuracy overall, but can not be directly applied to other datasets. We observed lower accuracy on Modality and Organ questions in VQA-RAD, which we attribute to the diversity of question and answer phrasing in these VQA-RAD question types.

\section{Attention Visualization}
\begin{figure*}[!h]
\centering
\includegraphics[width=\textwidth]{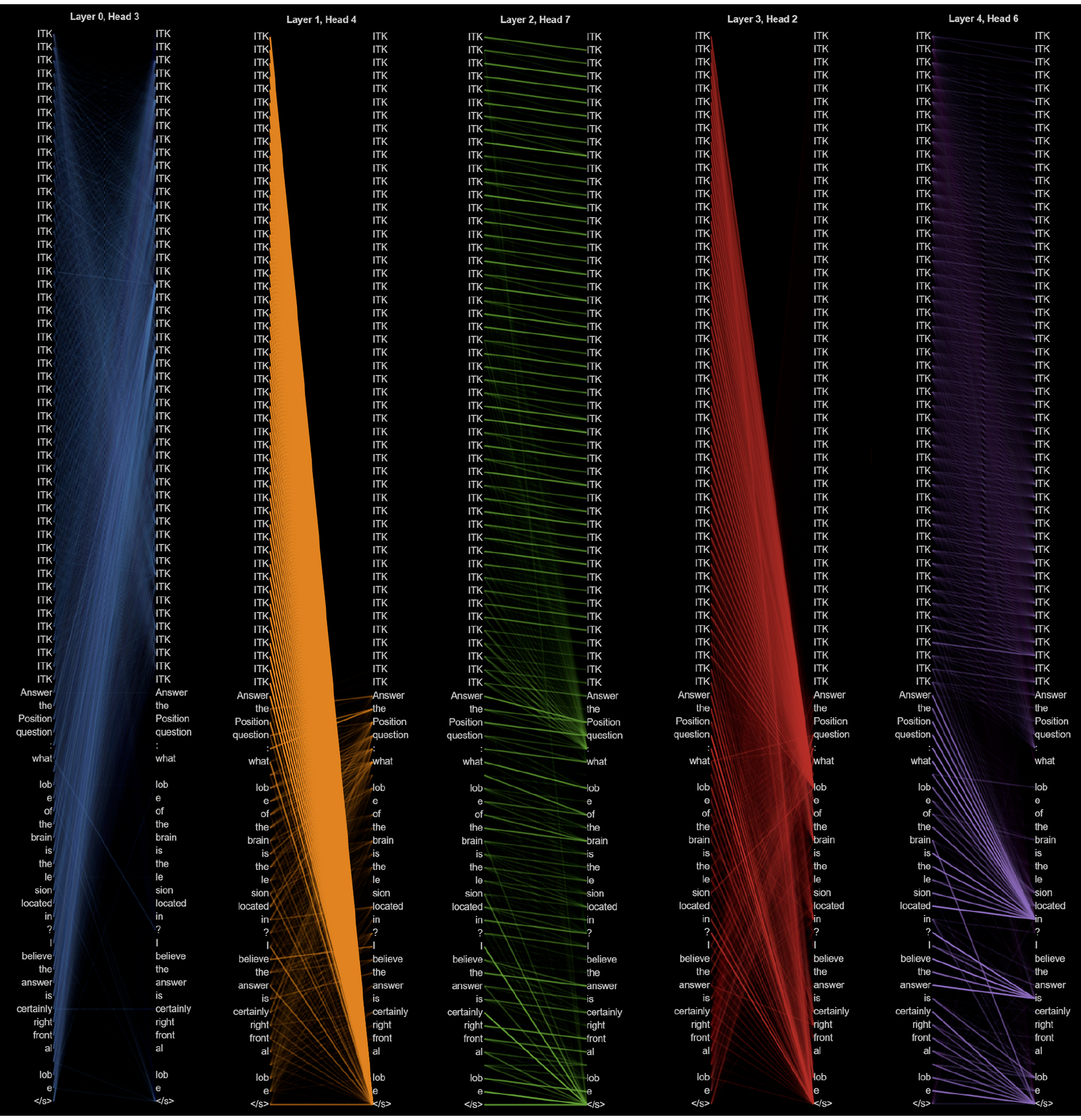}
\vspace{-3cm}

  \caption{Selected encoder self-attention visualizations across different encoder layers and attention heads. ITK represents an image-token from the input image. }
  \label{fig:self-attention}
\end{figure*}

Transformer-based models utilize attention to calculate dependencies between inputs which may be important for prediction. Since our MPR model uses a pretrained T5 encoder to combine features from several sources, a visualization of its attention scores may indicate which parts of the image contribute to its answers. Figure \ref{fig:self-attention} illustrates the encoder self-attention in different attention heads and layers of our model when asked a challenging position question. The results suggest that some attention modules may attend to the entire input image (Layer 1, Head 4),  whereas others may look for local dependencies by attending to adjacent tokens (Layer 2, Head 7).

\begin{figure*}[!h]
\centering
\includegraphics[width=\textwidth]{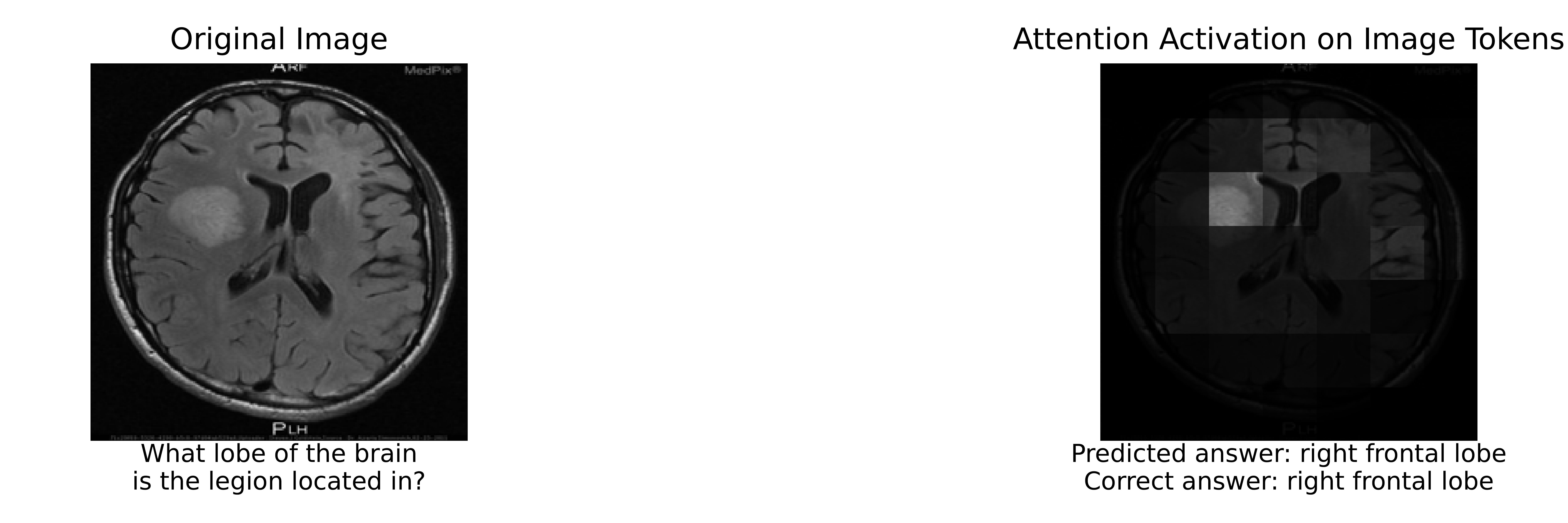}
\includegraphics[width=\textwidth]{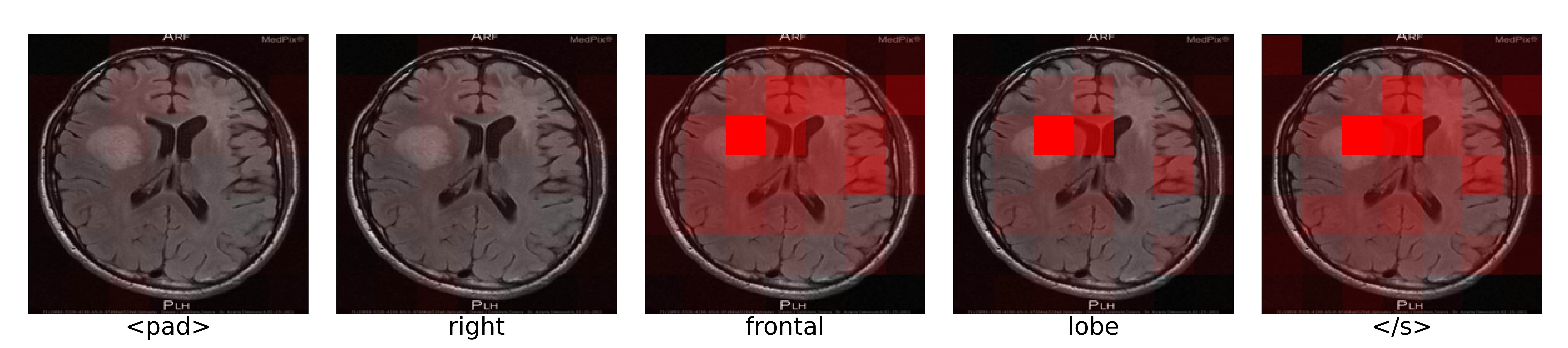}
  \caption{\textit{Top}: The original image and question input to the model, followed by the average attention scores for each image patch, with darker patches corresponding to lower scores. \textit{Bottom}: When predicting each word in the answer span, which input image regions are attended to.}
  \label{fig:cross_attentions}
\end{figure*}

In addition to encoder self-attention, cross attention in encoder-decoder transformer architectures may also illustrate which tokens from the input prompt contribute the most when generating answer tokens. Since the prompt to our model consists of image tokens, we visualize which image regions have the highest attention scores in \ref{fig:cross_attentions}. 

Existing work has shown the effectiveness of using retrieval from a data store to rapidly adapt language models to new domains \cite{khandelwal2019generalization}. KNN LMs uses a blending parameter $\lambda \in [0,1]$  to control the influence of retrieved information towards prediction:
\begin{align}
     \lambda \text{pkNN}(y|x) + (1-\lambda) \text{pLM}(y|x)
\end{align}

\begin{figure*}[!h]
\centering
\includegraphics[width=\textwidth]{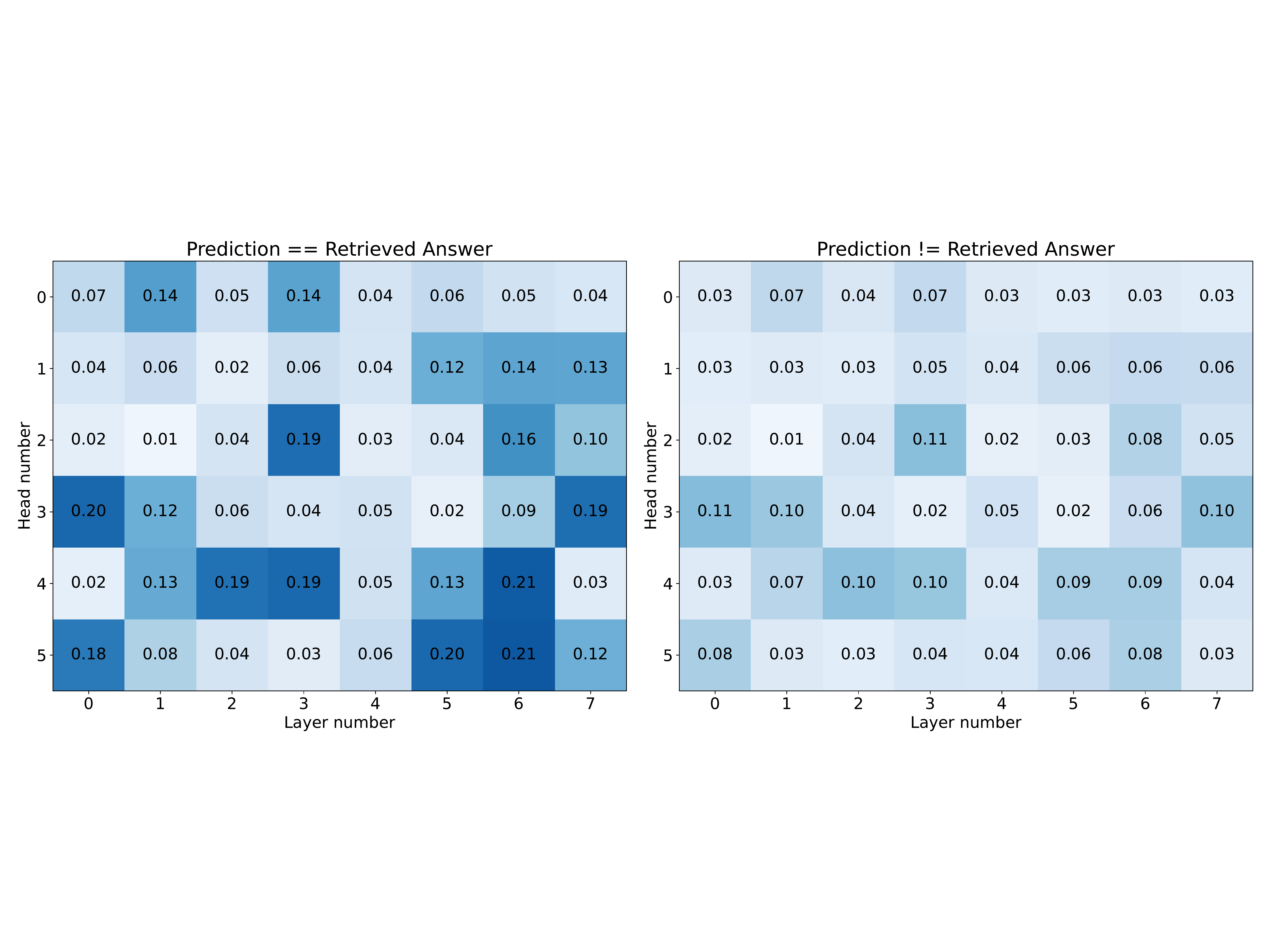}
\vspace{-3cm}
  \caption{Average attention score to the retrieval prompt for each attention module in our T5 encoder. When the model predicts the retrieved answer, attention scores to the retrieved information is significantly higher. }
  \label{fig:retrieved_answer_attention.pdf}
\end{figure*}

This method assumes the availability of a language model $\text{pLM}(y|x)$ and a retrieval model $\text{pkNN}(y|x) $which can predict the next vocabulary token $y$ given context $x$. However, given our retrieval set which consists of variable length answers and image data, it is difficult to estimate $\text{pkNN}(y|x)$ directly from our data store. Consequently, we augment our model input with retrieval prompts to allow for the implicit learning of retrieval reliance in an end-to-end manner. Figure \ref{fig:retrieved_answer_attention.pdf} shows the average cross attention scores to the retrieval portion of the prompt when evaluating on test data. The results demonstrate that when the model prediction matches the retrieved answer, the attention scores to the corresponding prompt section are significantly higher. Based on this observation, we believe the model has learned how to weigh the retrieved information through end-to-end training.


\end{document}